\documentclass[10pt,twocolumn,letterpaper]{article}

\usepackage{cvpr}             
\usepackage{graphicx}
\usepackage{amsmath}
\usepackage{amssymb}
\usepackage{booktabs}
\usepackage{mathtools, cuted}
\usepackage{mathrsfs}
\usepackage{breqn}

\usepackage{array}
\usepackage{amsfonts}
\usepackage[margin=1.0in]{geometry}
\usepackage{tabularx}
\usepackage{multirow}
\usepackage{url}
\usepackage{makecell}
\newcolumntype{P}[1]{>{\centering\arraybackslash}p{#1}}
\usepackage{caption}
\usepackage{subcaption}
\usepackage[pagebackref,breaklinks,colorlinks]{hyperref}
\usepackage[capitalize]{cleveref}
\crefname{section}{Sec.}{Secs.}
\Crefname{section}{Section}{Sections}
\Crefname{table}{Table}{Tables}
\crefname{table}{Tab.}{Tabs.}

\usepackage{wrapfig}
\usepackage{mathtools}
\usepackage{multirow}
\usepackage{booktabs}
\usepackage{subcaption}

\begin{document}

\title{Deep Residual Axial Networks}

\author{Nazmul Shahadat, Anthony S.\ Maida\\
University of Louisiana at Lafayette\\
Lafayette LA 70504, USA\\
{ nazmul.ruet@gmail.com, maida@louisiana.edu}}
\maketitle

\begin{abstract}
 While convolutional neural networks (CNNs) demonstrate outstanding performance on computer vision tasks, their computational costs remain high. 
Several techniques are used to reduce these costs like reducing channel count, and using separable 
and depthwise separable convolutions. 
This paper reduces computational costs by introducing a novel architecture, axial CNNs, which replaces spatial 2D convolution operations with two consecutive depthwise separable 1D operations. 
The axial CNNs are predicated on the assumption that the dataset supports approximately separable convolution operations with little or no loss of training accuracy. Deep axial separable CNNs still suffer from gradient problems when training deep networks.  
   We modify the construction of axial separable CNNs with residual connections to improve the performance of deep axial architectures and introduce our final novel architecture namely residual axial networks (RANs).
   Extensive benchmark evaluation shows that RANs achieve at least 1\% higher performance with about $77\%$, $86\%$, $75\%$, and $34\%$ fewer parameters, and about $75\%$, $80\%$, $67\%$, and $26\%$ fewer flops than ResNets, wide ResNets, MobileNets, and SqueezeNexts on CIFAR benchmarks, SVHN, and Tiny ImageNet image classification datasets. Moreover, our proposed RANs  improve deep recursive residual networks performance with 94\% fewer parameters on the image super-resolution dataset.
\end{abstract}

\section{Introduction}

Convolutional neural networks (CNNs) have emerged as a core building block for computer vision tasks, including classification \cite{he2016deep,he2015delving,7761223}, object detection \cite{lin2017feature,dai2016r} and image super-resolution \cite{kim2016accurate,kim2016deeply,tai2017image}. To solve major vision tasks, the CNN-based state-of-the-art architectures, specifically, ResNets \cite{he2016deep}, wide ResNets \cite{zagoruyko2016wide}, GoogleNets \cite{khan2019analysis}, AlexNets \cite{krizhevsky2017imagenet}, Xception \cite{chollet2017xception}, ResNeXt \cite{orhan2019robustness}, and hypercomplex CNNs \cite{shahadat2023enhancing,shahadat2023deep} have emerged in recent years. 
A common trend is to build deeper \cite{he2016deep,he2016identity} or wider \cite{zagoruyko2016wide,khan2019analysis} networks to improve performance. 
However, increasing the depth or widening the network also increases its computational costs.

A variety of CNNs were introduced to deal with these costs. 
Residual bottleneck blocks use $1\times 1$ pointwise 2D convolutions to reduce and then increase the channel counts. 
As a result, the spatial 2D CNN processes fewer channels and reduces the model's computational costs. But these are not enough
as they use standard 2D convolutions which consumes high costs.
This cost reduction has not been analyzed for wider ResNets as the widening factor ($\alpha$) multiplies the channel counts, raising the costs exponentially. The wide ResNets also use standard 2D convolutions.

As the standard 2D convolution is the core layer type of many computer vision models \cite{he2016deep,zagoruyko2016wide,khan2019analysis,krizhevsky2017imagenet,chollet2017xception,orhan2019robustness,shahadat2023enhancing,shahadat2023deep}, and it consumes high costs, several modifications have been applied to reduce these costs. A depthwise separable convolution (DSC) convolves independently over each input channel to minimize the costs $d_{in}$ (number of input channels) times than the standard convolution operations. 
Although this depthwise separable concept was introduced in 2014 for neural networks, it has been used more for CNN-based computer vision models, for example, Xception networks \cite{chollet2017xception}, and MobileNets \cite{howard2017mobilenets}. 
Among these, MobileNet is an efficient, lightweight deep DSC for mobile-based vision tasks. 
It reduces the costs by a factor of 8 or 9 at only a small reduction in accuracy. 
But it requires two CNN layers ($k\times k$ DSC layer and $1\times 1$ pointwise standard CNN layer) to replace a standard 2D CNN layer. So, it increases the layer count. Moreover, the pointwise layer still uses standard 2D convolutions. 

SqueezeNext \cite{gholami2018squeezenext} also reduces costs and
is guided by SqueezeNet \cite{iandola2016squeezenet} and separable convolution (SC) (replace $k\times k$ 2D convolution using filters $k\times 1$ and $1\times k$). 
This SC idea reduces the cost from $k^2$ to $2k$. 
They also squeeze the layer (like SqueezeNet \cite{iandola2016squeezenet}) before applying SC which also reduces cost. 
These models still use standard 2D convolutions.

Our work revisits the designs of the deep building blocks to boost their performance further, reduce computational costs, and improve the model's inference speed. To achieve these, we propose our novel architecture, residual axial networks (RANs), obtained by applying 1D DSC operations along the height and width-axis instead of SCs in the residual block, the InceptionV3 \cite{szegedy2016rethinking}, and the SqueezeNext \cite{gholami2018squeezenext} block. 
We split the 2D spatial ($3 \times 3$) convolution operation into two consecutive 1D DSC operations. 
These 1D DSC operations are mapped to the height and width axis. As axial 1D DSC operations propagate information along one axis at a time, this modification reduces cost at least $w\cdot k$ times (explain below). 
Moreover, this RAN block does not increase layer counts as two 1D layers equal one 2D layer.

A simple axial architecture reduces costs but does not improve performance. This is because forward information flowing across the axial blocks degrades (diminishing feature reuse \cite{10.1007/978-3-319-46493-0_39}). To address this, we add residual connections to span the axial blocks. By using both modifications, our novel and effective RANs improve validation performance. The effectiveness of our proposed model is demonstrated experimentally on four image classification datasets and an image super-resolution dataset. Our assessments are based on parameter counts, FLOP counts (number of multiply-add operations), latency to process one image after training, and validation accuracy.

\section{Background and Related Work}
\subsection{Convolutional Neural Networks}

In a convolutional layer, the core building block is
a convolution operation using a trainable weight $W$ for 2D multichannel images applied to small neighborhoods to find input correlations. 
For an input image $X$ with height $h$, width $w$, and channel count $d_{in}$, the convolution operation operates on region ${(a,b)} \in X$ centered at pixel $(i,j)$ with spatial extent $k$. The output for this operation $C_{i,j}$ is,
\begin{equation} 
C_{i,j,n} = \sum_{\mathclap{(a, b, m) \in \mathcal{N}_{k\times k(i,j)}}} W_{a, b, m, n} X_{i+a-1,j+b-1,m}
\label{conv:1} 
\end{equation}
where m, n, and $\mathcal{N}_{k\times k}$ are the index for input channel $d_{in}$, the index for output channel $d_{out}$, and the neighborhood of pixel $(i,j)$ with spatial extent $k$ of size $k\times k\times d_{in}\times d_{out}$, and $W$ is the shared weights to calculate the output for all pixel positions $(i,j)$. The computational cost of this convolutional operation is,  
\begin{equation}
    \mathrm{Cost}_{\mathrm{Conv2D}} = h\cdot w\cdot d_{in}\cdot d_{out}\cdot k\cdot k
\label{convCost:1}  
\end{equation} 
where the computational cost depends multiplicatively on the kernel size $k\times k$, feature map size $h\times w$, number of input channels (input depth) $d_{in}$, and number of output channels (output depth) $d_{out}$.

\subsection{Residual Networks}
\label{resnets}
Residual networks (ResNets) are constructed using 2D convolutional layers linked by additive identity connections \cite{he2016identity} for vision tasks. They were introduced to address the problem of vanishing gradients found in standard deep CNNs.

\begin{figure*}
    \centering
     \begin{subfigure}[b]{0.22\textwidth}
        \centering
        \includegraphics[width=\textwidth,height=150pt]{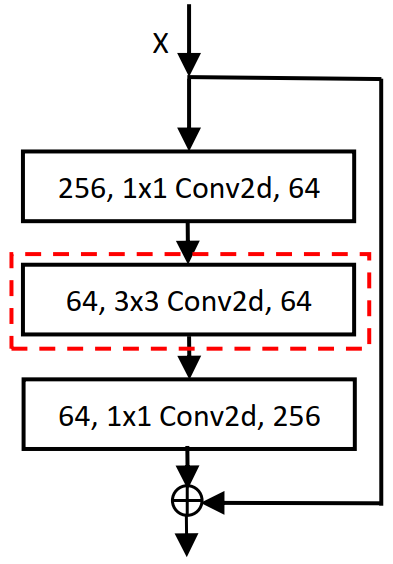}
        \caption{ResNet bottleneck block}
        \label{ResNet_block}
    \end{subfigure}
    \hfill
    \begin{subfigure}[b]{0.22\textwidth}
        \centering
        \includegraphics[width=\textwidth,height=200pt]{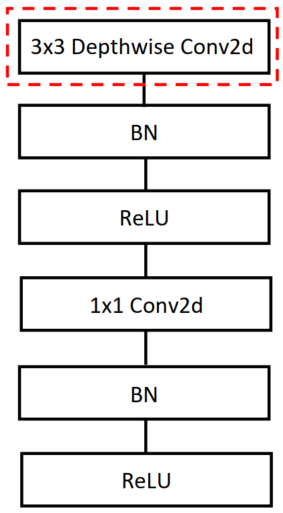}
        \caption{MobileNet block}
        \label{MobileNet}
    \end{subfigure}
    \hfill
    \begin{subfigure}[b]{0.23\textwidth}
        \centering
        \includegraphics[width=\textwidth,height=200pt]{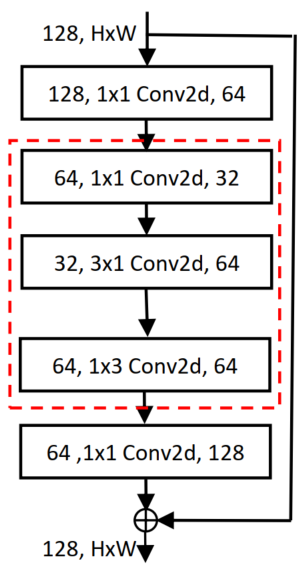}
        \caption{SqueezeNext block}
        \label{QNext}
    \end{subfigure}
    \hfill
    \begin{subfigure}[b]{0.23\textwidth}
        \centering
        \includegraphics[width=\textwidth,height=130pt]{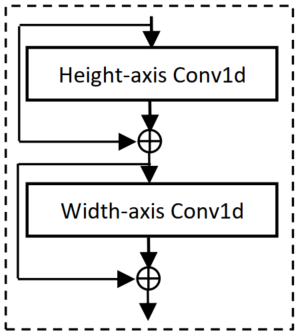}
        \caption{RAN block}
        \label{RAN}
    \end{subfigure}
    \hfill
\caption{Block types. ``bn'', ``ReLU'', and ``Conv2d'' stand for batch normalization, rectified linear unit, and 2D CNN, respectively. (a) Original Bottleneck modules found in ResNet \cite{he2016deep}, (b) original MobileNet block found in \cite{howard2017mobilenets}, (c) original SqueezeNext block found in \cite{gholami2018squeezenext}, and (d) novel Residual Axial Network (RAN) block used in our model.}
    \label{fig:proposedWithAttention}
\end{figure*}

The key architectural feature of ResNets is the residual block with identity mapping. 
Two kinds of residual blocks are used in residual networks, the basic block and the bottleneck block. We discuss the bottleneck block first. 
Figure \ref{ResNet_block} shows a bottleneck block for ResNets that is constructed using $1 \times 1$, $k \times k$, and $1 \times 1$ convolution layers, where the $1 \times1$ pointwise 2D CNN layers reduce and then increase the number of channels.
The $3 \times 3$ 2D CNN layer performs feature extraction. 
The identity shortcut connection is shown in the following equation,
\begin{equation}
y = \mathcal{F}(C_{k_1,k_m,k_1}(x,W))+x ,
\label{equ: bottleneck_conv} 
\end{equation}
where $\mathcal{F}, x, y,$ and $ W,$ represent residual function, input vector, output vector, and weight parameters. 
$C_{k_1,k_m,k_1}$ stands for the output of three convolution layers with kernels $1 \times 1$, $K_m\times K_m$, and $1 \times 1$, respectively. 
The computational cost of a $3\times 3$ spatial 2D CNN layer is given in Equation \ref{convCost:1}.
The cost of $1\times 1$ pointwise 2D CNN layers is given in Equation \ref{convCost:2}.
\begin{equation}
\mathrm{Cost}_{\mathrm{1x1Conv2D}} = h\cdot w\cdot d_{in}\cdot d_{out} 
\label{convCost:2}  
\end{equation} 
Hence, the computational cost of the residual bottleneck block is,
\begin{equation}
\mathrm{Cost}_{\mathrm{Bottle}} = h\cdot w\cdot d_{in}\cdot d_{out}\cdot k\cdot k + 2\cdot h\cdot w\cdot d_{in}\cdot d_{out} 
\label{convCost:3}  
\end{equation} 

In contrast to the bottleneck block, the basic architecture of ResNet is constructed with two $k\times k$ convolution layers where k is the size of the kernel and an identity shortcut connection is added to the end of these two layers. These operations can be expressed mathematically as,
\begin{equation}
y = \mathcal{F}(C_{k_m,k_n}(x,W))+x 
\label{equ: basic_conv} 
\end{equation}
where $C_{k_m,k_n}$ represents the output of two convolution layers with kernels $k_m$ and $k_n$ respectively. The computational cost of the residual basic block is,
\begin{equation}
\mathrm{Cost}_{\mathrm{Basic}} = 2\cdot h\cdot w\cdot d_{in}\cdot d_{out}\cdot k\cdot k 
\label{convCost:4}  
\end{equation} 
The performance of ResNets surpasses its learning speed, the number of learning parameters, the way of layer-wise representation, and memory mechanisms.

\subsection{Wide Residual Networks}
\label{wrns}

Wide ResNets \cite{zagoruyko2016wide,chen2020scaling} use fewer layers than standard ResNets but use higher channel counts (wide architectures) which 
compensate for their shallower architecture. 
Comparisons between shallow and deep networks have been studied in circuit complexity theory where shallow circuits require more components than deeper circuits. 
Inspired by this observation, \cite{he2016deep} proposed deeper networks with thinner architecture where a gradient goes through the layers. But the problem such networks face is that the residual block weights do not flow through the network layers. Because of this, the network may be forced to avoid learning during training. 
To address this, \cite{zagoruyko2016wide} proposed shallow but wide architectures and showed that widening the residual blocks improves the performance of residual networks compared to increasing their depth.  
The computational cost of this 2D convolutional operation is,  
\begin{equation}
\mathrm{Cost}_{\mathrm{Conv2D(WRNs)}} = h\cdot w\cdot d_{in}\cdot \alpha d_{out}\cdot k\cdot k ,
\label{convCost:5}  
\end{equation} 
where $\alpha$ is a widening factor.

\subsection{MobileNet Architectures}
\label{subsec:MobileNet}
Howard et al.\ developed a mobile-based shallower network for vision tasks depicted in Figure \ref{MobileNet}. They used DSCs because it helps to build lightweight networks. A pointwise $1\times 1$ convolution is used to combine the outputs of DSC \cite{howard2017mobilenets}. These two steps are performed in standard convolution in a single step. This DSC performs convolution 
per input channel and it can be defined as,
\begin{equation} 
C_{i,j,n} = \sum_{\mathclap{(a, b) \in \mathcal{N}_{k\times k(i,j)}}} W_{a, b, n} X_{i+a-1,j+b-1,n}
\label{DepthwiseConv} 
\end{equation}
where the $n_{th}$ channel of trainable weight $W$ is applied to the $n_{th}$ channel of input $x$ to produce the $n_{th}$ channel of the output feature map $C$. The computational cost of this 2-dimensional depthwise separable convolutional operation is,  
\begin{equation}
\mathrm{Cost}_{\mathrm{DWConv2D}} = h\cdot w\cdot d_{out}\cdot k\cdot k.
\label{convCost:6}  
\end{equation} 
And the pointwise $1\times 1$ convolution has a computational cost which is explained in Equation \ref{convCost:2}.
The computational cost of depthwise separable convolutions is,
\begin{equation}
\mathrm{Cost}_{\mathrm{DWConv2Ds}} = h\cdot w\cdot d_{out}\cdot k\cdot k+h\cdot w\cdot d_{in}\cdot d_{out}
\label{convCost:8}  
\end{equation}
which is the sum of the computational costs of depthwise (Equation \ref{convCost:6}) and pointwise (Equation \ref{convCost:2}) convolutions.

\subsection{SqueezeNext Architecture}
\label{QueezeNext}
A CNN with fewer input and output channels requires fewer trainable parameters, less cross-server communication for distributed training, lower bandwidth to export, and is easier to deploy on field-programmable gate arrays (FPGAs) with limited memory \cite{iandola2016squeezenet}. 
To achieve these advantages, Iandola et al.\ proposed SqueezeNet (SNet) where they squeeze the input channels to reduce the number of filters \cite{iandola2016squeezenet}. The computational cost of this SqueezeNet spatial conv2d operation is,
\begin{equation}
\mathrm{Cost}_{\mathrm{Conv2D(SNet)}} = h\cdot w\cdot ds_{in}\cdot d_{out}\cdot k\cdot k
\label{convCost:9}  
\end{equation} 
where $ds_{in}$ is the squeezed input channels. Gholami et al. further reduce this cost by applying separable convolutions ($3\times 1$ and $1\times 3$ Conv2d) 
instead of a spatial $3\times 3$ convolution (Conv2d) and they called it SqueezeNext \cite{gholami2018squeezenext}. 
It reduces the cost once again compared to SqueezeNet and the new cost of the conv2d layer is,
\begin{equation}
\begin{split}
\mathrm{Cost}_{\mathrm{SqNext}}
& = h\cdot d_{in}\cdot d_{out}\cdot k\cdot k + w\cdot d_{in}\cdot d_{out}\cdot k\cdot k \\
& = 2\cdot h\cdot d_{in}\cdot d_{out}\cdot k\cdot k.
\end{split}
\label{convCost:10}  
\end{equation} 
As height, $h$ equals width $w$ for computer vision tasks. The SqueezeNext block is depicted in Figure \ref{QNext}. 
SqueezeNext performed better than VGG-19 \cite{simonyan2014very}, AlexNet \cite{krizhevsky2017imagenet}, SqueezeNet \cite{iandola2016squeezenet}, and MobileNet \cite{howard2017mobilenets}.

\subsection{Recursive Residual Networks}
Image super-resolution (SR) is the process of generating a high-resolution (HR) image from a low-resolution (LR) image. It is also known as single image super-resolution (SISR). Convolution-based recursive neural networks have been used on SISR \cite{dong2015image,kim2016accurate,kim2016deeply,tai2017image}, where recursive networks learn detailed and structured information about an image.  
Kim et al. introduce two deep CNNs for SR by stacking weight layers \cite{kim2016deeply,kim2016accurate}. Among them, \cite{kim2016deeply} uses a chain structure recursive layer along with skip-connections to control the model parameters and improve the performance. Deep SR models \cite{kim2016deeply,mao2016image,kim2016accurate} demand large parameter counts and more storage. 

To address these issues, deep recursive residual networks (DRRNs) were proposed as a very deep network structure, 
which achieves better performance with fewer parameters \cite{tai2017image}. 
It includes both local and global residual learning (GRL), where GRL is used in the identity branch to estimate the residual image from the input and output of the network. GRL might face degradation problems for deeper networks. 
To address this, local residual learning (LRL) has been used which carries rich image details to deeper layers and helps gradient flow. 
The DRRN also used recursive learning of residual units to keep the model more  compact. Several recursive blocks (B) have been stacked, followed by a CNN layer which is used to reconstruct the residual between the LR and HR images. 
Each of these residual blocks decomposes into the number of residual units (U). 
The number of $B$ and $U$ is responsible for defining network depth. The depth of DRRN $d$ is calculated as,
\begin{equation}
    d=(1+2\times U) \times B + 1
    \label{equ:DRRN}
\end{equation}
Recursive block definition, DRRN formulation, and the loss function of DRRN are defined in \cite{tai2017image}. The computational cost of each unit $U$ will be the same as in equation \ref{convCost:4}.

\section{Proposed Residual Axial Networks}
The 2D CNN is highly performant with the help of several state-of-the-art architectures like, 
ResNets \cite{he2016deep}, wide ResNets \cite{wu2019wider}, scaling wide ResNets \cite{zagoruyko2016wide}, 
MobileNets \cite{howard2017mobilenets}, SqueezeNets \cite{iandola2016squeezenet}, SqueezeNexts \cite{gholami2018squeezenext}, 
and deep recursive residual networks (DRRNs) \cite{tai2017image}  on image classification and image super-resolution datasets. 
The residual bottleneck block makes the networks thinner; still, 
the cost efficiency of these blocks can be improved.
The cost of 2D convolution, residual bottleneck, and basic blocks is calculated in Equations \ref{convCost:1}, \ref{convCost:3}, and \ref{convCost:4}, respectively.
The 2D convolution operation given in Equation \ref{conv:1} uses a $k\times k$ filter for the input $X\in h\times w\times d_{in}$. 
Equation \ref{convCost:1} gives the costs of 2D convolution in the residual blocks. 
SC is used to reduce these costs in InceptionNetV3 \cite{wang2019pulmonary}, SqueezeNext \cite{gholami2018squeezenext}, and axial-deeplab \cite{wang2020axial}. They decomposed this $k\times k$ convolution into two separable convolutions with $k\times 1$ and $1\times k$ sized filters. This decomposition effectively reduces the number of parameters from $h\times w\times d_{in}\times d_{out}\times k\times k$ to $2\times h\times w\times d_{in}\times d_{out}\times k$. Their decomposed convolution with spatial extent $k\times 1$ is defined as,
\begin{equation} 
C_{i,j,n} = \sum_{\mathclap{(a, b, m) \in \mathcal{N}_{k\times 1(i,j)}}} W_{a, b, m, n} X_{i+a-1,j+b-1,m}
\label{equ:proposed1} 
\end{equation}
where, $m$, and $n$ are the indices for input channel $d_{in}$, and for output channel $d_{out}$. Also, $\mathcal{N}_k \in \mathbb{R}^{k\times 1\times d_{in}}$ is the neighborhood of pixel $(i,j)$ with spatial extent $k\times 1$ and $W \in \mathbb{R}^{k\times 1\times d_{out} \times d_{in}}$ is the shared weights that are for calculating output for all pixel positions $(i,j)$.  
For spatial extent $1\times k$ is defined as,
\begin{equation} 
C_{O(i,j,n)} = \sum_{\mathclap{(a, b, m) \in \mathcal{N}_{1\times k(i,j)}}} W_{a, b, m, n} C_{i+a-1,j+b-1,m}
\label{equ:proposed2} 
\end{equation}
where, $\mathcal{N}_k \in \mathbb{R}^{1\times k\times d_{in}}$ is the neighborhood of pixel $(i,j)$ with extent $1\times k$ 
and $W \in \mathbb{R}^{1\times k\times d_{out} \times d_{in}}$ is the shared weights that calculate output for all pixel positions $(i,j)$.  

Their cost efficiency can be improved
as they still use 2D convolutions with spatial extent filters for 2D inputs (even though one dimension has a size equal to 1). 
To reduce these costs further, we propose a novel residual axial network (RAN) to replace any 2D spatial convolutional layer in a network. 
We replace 2D convolution operations (conv2D) of any block by using two 1D DSC operations with filters $k$. 
To apply 1D DSC in 2D input size of $h\times w$, we split the inputs into the height and width axes. 
Each 1D convolution layer applies to each input axis. 
The 1D DSC operation is defined as,
\begin{equation} 
C_{O(i,n)} = \sum_{\mathclap{a \in \mathcal{N}_{k(i)}}} W_{a,n} X_{i+a-1,n}
\label{equ:proposed3} 
\end{equation}
where $\mathcal{N}_k \in \mathbb{R}^{k\times d_{in}}$ is the neighborhood of pixel $i$ with extent $k$ and $W \in \mathbb{R}^{k\times d_{out} \times d_{in}}$ 
is the shared weights that calculate output for all pixel positions $i$. 
In Equation \ref{equ:proposed3}, the $n_{th}$ channel of trainable weight $W$ is applied to the $n_{th}$ channel of input $X$ to produce the $n_{th}$ channel of the output feature map $C_O$.
The computational cost of this 1D DSC operation is calculated as
\begin{equation}
\mathrm{Cost}_{\mathrm{Conv1D}} = h\cdot d_{out}\cdot k.
\label{equ:proposed4}  
\end{equation} 
As our RAN block has two layers of 1D convolutions, the cost of this block is $2\cdot h\cdot d_{out}\cdot k$. Also, each 1D DSC operation has a residual connection to avoid vanishing gradients. 
Hence, our proposed novel architecture factorizes 2D convolution into two consecutive 1D DSCs along with residual connections depicted in Figure \ref{RAN}. 

To construct residual blocks, we replace each 2D convolution layer from the residual basic and bottleneck blocks using our RAN block. 
The spatial 2D convolution in the residual bottleneck block (red marked in Figure \ref{ResNet_block}) is replaced by the RAN block to construct our proposed RAN-based residual bottleneck block. In the same way, the proposed RAN-based basic block architecture is constructed. To compare our proposed parameter-efficient RAN block with 2D DSC, the depthwise 2D spatial convolution (red marked in Figure \ref{MobileNet}) of MobileNet is replaced by our proposed RAN block and constructed new RAN-based MobileNet. We also compare our proposed RAN block for separable convolution by replacing the red-marked area in Figure \ref{QNext} of SqueezeNext with the RAN block. The RAN block in Figure \ref{RAN} is applied to other 2D convolution-based networks, for example, wide residual networks (to make our proposed wide RANs) and deep recursive residual networks (to make RARNets depicted in Figure \ref{fig:RARNet}), to check the effectiveness of our proposed method in all possible ways.

\section{How RANs are Cost Effective}
\label{Why_RANs}
This section compares the computational costs of different aspects of CNN layers to the networks. The costs of standard convolutional operation, residual bottleneck and basic blocks, wide residual convolutional operation, MobileNet block, SqueezeNet, and SqueezeNext block are calculated in Equations \ref{convCost:1}, \ref{convCost:3}, \ref{convCost:4}, \ref{convCost:5}, \ref{convCost:8}, \ref{convCost:6}, and \ref{convCost:9}. The computational costs of our proposed RAN block compared with the standard 2D convolutional operation, we get a reduction in the computation of:
\begin{equation}
\begin{split}
\mathrm{Cost}_R & = \dfrac{\mathrm{Cost~of~2D~Convolution}}{2\cdot \mathrm{Cost~of~1D~Convolution}} \\
& = \dfrac{h\cdot w\cdot d_{in}\cdot d_{out}\cdot k^2} {2\cdot h\cdot d_{out}\cdot k} = \dfrac{w\cdot d_{in}\cdot k}{2}
\label{Equ:comparison1}
\end{split}
\end{equation} 
where $\mathrm{Cost}_R$ is the cost reduction ratio where the RAN block reduces costs $(w\cdot d_{in}\cdot k)\mathbin{/}2$ times than the original standard 2D convolution. 
Hence, the RAN block can be used as a replacement for any networks where 2D convolutional layers are used. We apply this block as a replacement of 2D CNN in residual networks \cite{he2016deep} specifically residual basic block (replacing the two 2D CNN layers), and  bottleneck block (replacing the only spatial 2D CNN layer) and construct RAN-based ResNet blocks. These RAN-based ResNet basic blocks reduce the costs of:
\begin{equation}
\begin{split}
\mathrm{Cost}_{R} & = \dfrac{2\cdot \mathrm{Cost~of~2D~Convolution}}{2\cdot \mathrm{Cost~of~1D~Convolution}} \\
 & = \dfrac{2\cdot h\cdot w\cdot d_{in}\cdot d_{out}\cdot k\cdot k} {2\cdot h\cdot d_{out}\cdot k} 
 = w\cdot d_{in}\cdot k
\label{Equ:comparison2}
\end{split}
\end{equation} 
where $\mathrm{Cost}_{R}$ is the cost reduction ratio where the RAN-based ResNet basic block reduces costs $(w\cdot d_{in}\cdot k)$ times than the original ResNet basic block. For the ResNet bottleneck block, the RAN-based bottleneck block reduces costs similar to the cost reduction in Equation \ref{Equ:comparison1}.

Now, we compare the cost-effectiveness of MobileNet and SqueezeNext architectures with our proposed RAN block-based MobileNet and SqueezeNext architectures. Our proposed RAN-based MobileNet block performs a reduction in the costs of:
\begin{equation}
\begin{split}
\mathrm{Cost}_R & = \dfrac{\mathrm{Cost_{DWConv2D}~in~MobileNetV1}}{2\cdot \mathrm{Cost~of~1D~Convolution}} \\
& = \dfrac{h\cdot w\cdot d_{out}\cdot k\cdot k}{2\cdot h\cdot d_{out}\cdot k} = \dfrac{w\cdot k}{2}
\label{Equ:comparison4}
\end{split}
\end{equation} 
where $d_{in}$ is $1$ for the original and proposed spatial convolutions as both networks use depthwise separable convolutions. There is a huge $(w\cdot k)\mathbin{/}2$ ($75\%$ reduction for CIFAR data in Table \ref{tab_resultCifar}) reduction for this MobileNet architecture. For SqueezeNext architectures, our RAN-based SqueezeNext reduces the computational costs of:
\begin{equation}
\begin{split}
\mathrm{Cost}_R &= \dfrac{\mathrm{PW1x1Conv2D ~+~2\cdot kx1Conv2D}}{2\cdot \mathrm{Cost~of~1D~Convolution}} \\
& = \dfrac{h\cdot w\cdot d_{in}\cdot d_{out}+2\cdot h\cdot d_{in}\cdot d_{out}\cdot k\cdot k}{2\cdot h\cdot d_{out}\cdot k} \\
& = \dfrac{w\cdot d_{in}}{2\cdot k} + d_{in}\cdot k
\label{Equ:comparison5}
\end{split}
\end{equation} 
where $\mathrm{Cost}_R$ is the cost reduction ratio of the original SqueezeNext and our RAN-based SqueezeNext blocks. A pointwise $1\times 1$ ($\mathrm{PW1x1Conv2D}$), two separable 2D convolutions with filters $k\times 1$, and $1\times k$ in SqueezeNext block are replaced by our proposed RAN block. 
Our RAN-based SqueezeNext block takes $(w\cdot d_{in})\mathbin{/}(2\cdot k)+(d_{in}\cdot k)$ times fewer computing from separable convolutional operations in the original SqueezeNext block. 
These formulas show that our proposed RAN block is parameter-efficient and cost-effective in replacing any 2D convolution for computer vision tasks.

\section{Experimental Analysis}
\label{Experimental_Analysis}
We present experimental results on four image classification datasets and an image super-resolution dataset. Our experiments  evaluate the original ResNets, wide ResNets, RAN-ResNets, wide RANs, MobileNet architectures, SqueezeNext architectures, RAN-based MobileNet and RAN-based SqueezeNext architectures, DRRNs, and RARNets. 
We compare our proposed RAN-based network's with the original ResNets, as these original networks used 2D CNN layers. 
Our comparisons use parameter counts, FLOPS, latency, and validation performance.

\subsection{Method: Convolutional Networks}

To explore scalability, we compare our proposed RANs and baseline models on four datasets: CIFAR-10 and CIFAR-100 benchmarks \cite{krizhevsky2009learning}, Street View House Number (SVHN) \cite{netzer2011reading}, and Tiny ImageNet datasets \cite{Le2015TinyIV}. The CIFAR benchmarks have 10 and 100 distinct classes and 60,000 color images 
of size $32\times32$. We perform data normalization using per-channel mean and standard deviation. In preprocessing, we do the horizontal flips and randomly crop after padding with four pixels on each side of the image. The SVHN and Tiny ImageNet datasets contain 600,000 images of size $32\times32$ with ten classes and 110,000 images of 200 distinct classes downsized to $64\times64$ colored images, respectively. Our only preprocessing is mean/std normalization for both datasets. All models were run using the stochastic gradient descent optimizer and linearly warmed-up learning for ten epochs from zero to 0.1 and then used cosine learning scheduling from epochs 11 to 150. 

\subsubsection{Residual Networks}
\label{method:ResNet}
ResNets and our proposed RAN-based ResNets were trained using similar designs (same hyperparameters and output channel counts). 
As our main concern was to reduce parameter counts of the residual bottleneck block, we implemented all baselines and the 
proposed architecture using only bottleneck blocks. 
The output channels of bottleneck groups are $120, 240, 480,$ and $960$ for all  networks. 
This experiment analyzes $26, 35, 50, 101,$ and $152$-layer architectures with the bottleneck block 
multipliers ``$[1, 2, 4, 1]$'', ``$[2, 3, 4, 2]$'', ``$[3, 4, 6, 3]$'', ``$[3, 4, 23, 3]$'', and ``$[3, 8, 36, 3]$'', respectively. 
All models were trained using batch sizes of 128 for all datasets except the 101 and 152-layer architectures of the Tiny ImageNet dataset. 
Batch size was 64 for these two architectures.

\subsubsection{Wide Residual Networks}
\label{method:WideResNet}
Section \ref{method:ResNet} explained the method for deeper networks. 
This section describes methods for wide but shallow networks. 
To assess the widening factor on our proposed RANs, we increase the width of our RANs by factorizing the number of output channels for shallow networks like \cite{zagoruyko2016wide}. Like the original wide residual networks (WRNs) \cite{zagoruyko2016wide}, we analyzed our proposed 26-layer bottleneck block of RANs with a widening factor, $k = 2, 4, 6, 8,$ and $10$. We multiplied the number of output channels of RANs with $k$ to obtain wide RANs. 
We trained with the same optimizer and hyperparameters used in Section \ref{method:ResNet}.

\subsubsection{MobileNet Architectures}
\label{method:MobileNet}
MobileNet and RAN-based MobileNet architectures use the hyperparameters and number of output channels similar to Table $1$ in \cite{howard2017mobilenets}. For the MobileNet architectures, we also use a $0.045$ initial learning rate decaying by 0.98 per epoch. Moreover, the standard RMSProp optimizer with decay and momentum is set to $0.9$. Unlike original MobileNets \cite{howard2017mobilenets}, we trained the original MobileNet and our proposed RAN-based MobileNet architectures using a batch size of 128.

\subsubsection{SqueezeNext Architectures}
\label{method:SqueezeNext}
For a fair comparison, we use similar hyperparameters to the original SqueezeNext \cite{gholami2018squeezenext}. 
The output channels of our RAN-based SqueezeNext groups are similar to the original SqueezeNext networks. This experiment analyzes 23-layer architectures with the block multipliers ``[6, 6, 8, 1]”. We analyze two 23-layer architectures, ``SqNxt-23-1x'', and ``SqNxt-23-2x'' where the channel widening factors are $1$, and $2$, respectively. All models were trained using batch sizes of 128 for CIFAR10, CIFAR100, and SVHN datasets.

\begin{figure}[ht]
    \centering
    \includegraphics[width=0.40\textwidth,height=130pt]{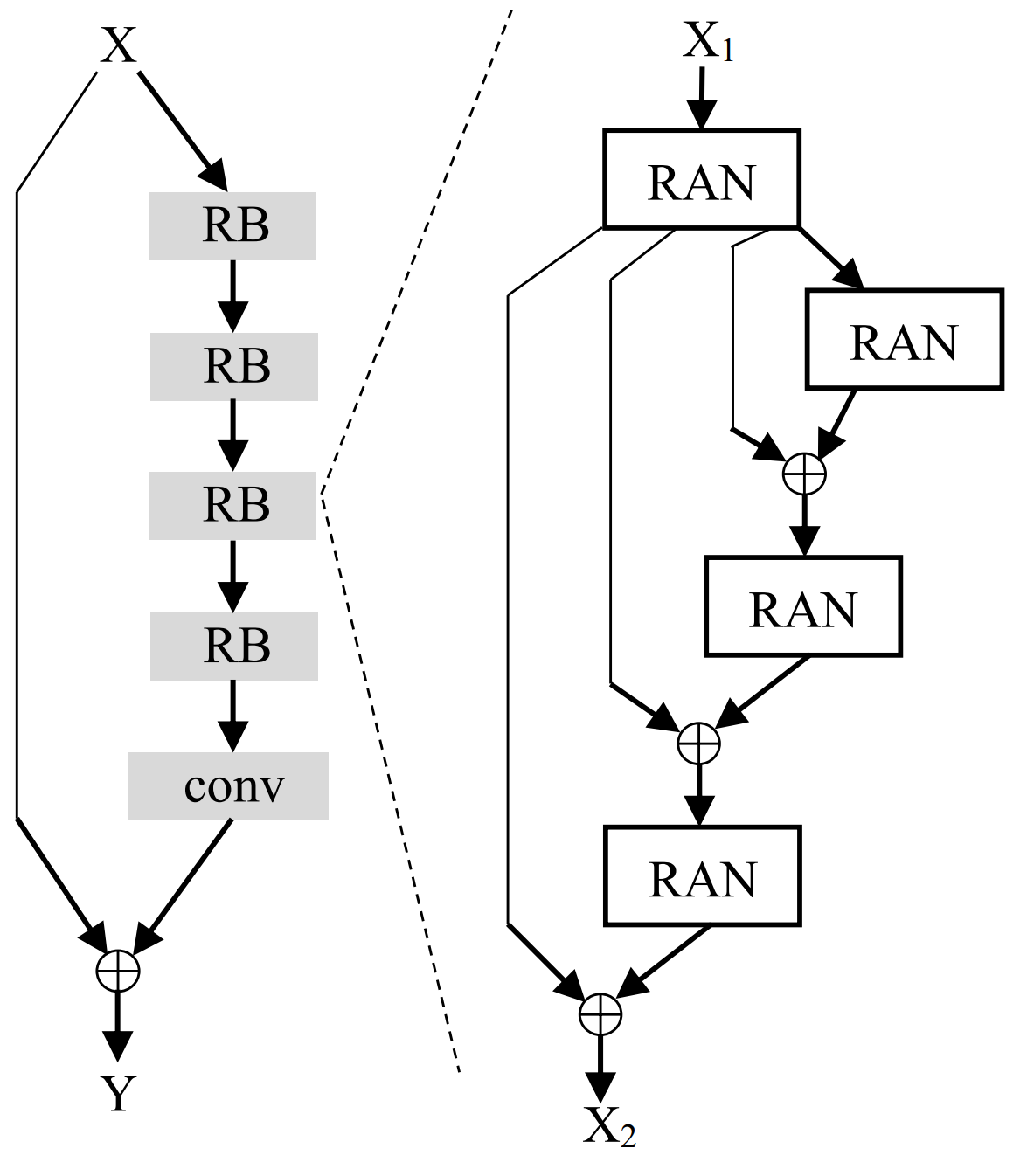}
    \caption{Recursive axial residual network (RARNet) architecture with $B=4$ and $U=3$. Here, the ``RB'' layer refers to a recursive block.}
    \label{fig:RARNet}
\end{figure}

\subsubsection{Recursive Networks}
This experiment compares the cost and performance of our novel RARNet with the DRRN on the super-resolution tasks.
The RARNet is built by replacing
the residual unit $U$ with a RAN block described in Equation \ref{equ:proposed3} and in Figure \ref{RAN}. 
These modifications form a new network, a recursive axial residual network whose depth $d$ is given by
\begin{equation}
    d=(1+U_{RAN})\times B+1 .
    \label{equ:RARNet}
\end{equation}
As two 1D layers are equivalent to one 2D layer and we replace each residual unit with a RAN unit (see Equation \ref{equ:proposed3}). Hence, we rewrite Equation \ref{equ:DRRN} to Equation \ref{equ:RARNet} by removing the multiplier for the residual unit. The proposed RARNet with four RB blocks is shown on the left, and an RB block is expanded on the right in Figure \ref{fig:RARNet}.

We trained our proposed RARNet using 291 images dataset \cite{yang2010image} and tested using the Set5 dataset \cite{bevilacqua2012low}. We also use different scales ($\times2$, $ \times3$, and $\times4$) in training and testing images. We used similar data augmentation, training hyperparameters, and implementation details like \cite{tai2017image}.

\subsection{Results Analysis}
\subsubsection{Residual Networks}
Table \ref{tab_resultCifar} summarizes the classification results of the original ResNets and our proposed RANs on the four datasets. We tested shallow and deeper networks by implementing $26, 35, 50, 101,$ and $152$-layer architectures. These architectures compare performance to check the effectiveness of our proposed methods for shallow and deep networks. Our proposed method is compared with original ResNets in terms of parameter count, FLOPS count, latency, and validation accuracy on the four datasets. 

The $26, 35, 50, 101,$ and $152$-layer architectures reduce by $77\%$, $76.9\%$, $76.7\%$, $76.6\%$, and $76.5\%$ trainable parameters respectively in comparison to the baseline networks. In addition to parameter reduction, our proposed method requires $15$ to $36$ percent fewer FLOPS for all analyzed architectures. Also, the validation performance improvement is significantly noticeable for all datasets in Tables \ref{tab_resultCifar} and \ref{tab_resultImageNet}. 
The latency of our proposed models is also lower than the original networks. 
Also, the deeper networks perform better than the shallow networks, demonstrating ``the deeper, the better'' in classification.

\begin{table*}[]
\centering
\begin{tabular}{|l|cc|cc|cc|cc|cc|cc|}
\hline
\multirow{2}{*}{\shortstack{Models}} & \multicolumn{2}{c|}{Params (M)} & \multicolumn{2}{c|}{FLOPs} & \multicolumn{2}{c|}{Latency (ms)} & \multicolumn{2}{c|}{CIFAR10} & \multicolumn{2}{c|}{CIFAR100} & \multicolumn{2}{c|}{SVHN} \\ \cline{2-13} 

& \multicolumn{1}{c|}{Orig} & Our&\multicolumn{1}{c|}{Orig} & Our&\multicolumn{1}{c|}{Orig} & Our& \multicolumn{1}{c|}{Orig} & Our & \multicolumn{1}{c|}{Orig} & Our & \multicolumn{1}{c|}{Orig} & Our \\ \hline

ResNet-26 & \multicolumn{1}{c|}{40.9} & \textbf{9.4} & \multicolumn{1}{c|}{2.56G} & \textbf{0.68G} & \multicolumn{1}{c|}{0.66} & \textbf{0.52} & \multicolumn{1}{c|}{94.68} & \textbf{96.08} & \multicolumn{1}{c|}{78.21} & \textbf{79.66} & \multicolumn{1}{c|}{96.04} & \textbf{97.83} \\ \hline

ResNet-35 & \multicolumn{1}{c|}{58.1} & \textbf{13.4} & \multicolumn{1}{c|}{3.26G} & \textbf{0.86G} & \multicolumn{1}{c|}{0.66} & \textbf{0.60} & \multicolumn{1}{c|}{94.95} & \textbf{96.15} & \multicolumn{1}{c|}{78.72} & \textbf{80.38} & \multicolumn{1}{c|}{95.74} & \textbf{97.50} \\ \hline

ResNet-50 & \multicolumn{1}{c|}{82.5} & \textbf{19.2} & \multicolumn{1}{c|}{4.58G} & \textbf{1.16G} & \multicolumn{1}{c|}{1.11} & \textbf{0.73} & \multicolumn{1}{c|}{95.08} & \textbf{96.25} & \multicolumn{1}{c|}{78.95} & \textbf{81.29} & \multicolumn{1}{c|}{95.76} & \textbf{97.32} \\ \hline

ResNet101 & \multicolumn{1}{c|}{149} & \textbf{34.8} & \multicolumn{1}{c|}{8.8G} & \textbf{2.18G} & \multicolumn{1}{c|}{1.68} & \textbf{1.28} & \multicolumn{1}{c|}{95.36} & \textbf{96.27} & \multicolumn{1}{c|}{78.80} & \textbf{80.88} & \multicolumn{1}{c|}{96.29} & \textbf{97.29} \\ \hline

ResNet152 & \multicolumn{1}{c|}{204} & \textbf{47.9} & \multicolumn{1}{c|}{13.1G} & \textbf{3.2G} & \multicolumn{1}{c|}{2.36} & \textbf{1.80} & \multicolumn{1}{c|}{95.36} & \textbf{96.37} & \multicolumn{1}{c|}{79.85} & \textbf{80.94} & \multicolumn{1}{c|}{96.35} & \textbf{97.38} \\ \hline

MobileNet & \multicolumn{1}{c|}{3.2} & \textbf{0.8} & \multicolumn{1}{c|}{ 12M} & \textbf{4M} & \multicolumn{1}{c|}{0.18} & \textbf{0.18} & \multicolumn{1}{c|}{87.87} & \textbf{93.34} & \multicolumn{1}{c|}{60.64} & \textbf{61.1} & \multicolumn{1}{c|}{94.23} & \textbf{94.53} \\ \hline

SqNxt23-1 & \multicolumn{1}{c|}{0.6} & \textbf{0.4} & \multicolumn{1}{c|}{ 59M} & \textbf{45M} & \multicolumn{1}{c|}{0.55} & \textbf{0.37} & \multicolumn{1}{c|}{92.30} & \textbf{93.34} & \multicolumn{1}{c|}{69.70} & \textbf{70.14} & \multicolumn{1}{c|}{95.88} & \textbf{97.13} \\ \hline

SqNxt23-2 & \multicolumn{1}{c|}{2.3} & \textbf{1.7} & \multicolumn{1}{c|}{ 226M} & \textbf{168M} & \multicolumn{1}{c|}{0.78} & \textbf{0.47} & \multicolumn{1}{c|}{93.38} & \textbf{94.91} & \multicolumn{1}{c|}{73.05} & \textbf{74.94} & \multicolumn{1}{c|}{96.06} & \textbf{97.40} \\ \hline

\end{tabular}
\caption{Image classification performance on the CIFAR benchmarks and SVHN datasets for different architectures. Here, ``Orig'', ``Our'', ``SqNxt23-1'', and ``SqNxt23-2'' are the original and our proposed network for corresponding models, and 23 layers SqueezeNext with widening factor 1 and 2.}
\label{tab_resultCifar}
\end{table*}

\subsubsection{Wide Residual Networks}
Table \ref{tab_resultCifar} shows ``the deeper, the better'' in vision classification for our proposed methods. To compare our proposed RANs with the original WRNs, we analyze our proposed method for different widening factors. Table \ref{tab_resultWRNs} shows an overall comparison among the original WRN-28-10 (28-layers with a widening factor of 10) and our proposed 26-layer networks with different widening factors ($2, 4, 6, 8,$ and $10$). Our proposed wide RANs (WRANs) show 4\% better performance with 86\% fewer parameters than the original WRN \cite{zagoruyko2016wide}. This table also demonstrates ``the wider, the better'' for our proposed WRANs.

\begin{table}[]
\centering
\begin{tabular}{|l|c|c|c|c|}
\hline
Models  & Params & FLOPs & Latency & Accu. \\ \hline
 ResNet-26 & 41.6M & 0.66G & 2.31ms & 57.21 \\ \cline{1-5} 
 
\textbf{RANs-26} &  \textbf{21.3M} & \textbf{0.56G} & \textbf{2.58ms} & \textbf{62.28} \\ \cline{1-5}

ResNet-35 & 58.5M & 0.86G & 2.85ms & 57.80 \\ \cline{1-5} 
\textbf{RANs-35} &  \textbf{31.3M} & \textbf{0.68G} & \textbf{3.0ms} & \textbf{59.31} \\ \cline{1-5} 

 ResNet-50 & 82.6M & 1.18G & 3.75ms & 59.06 \\ \cline{1-5} 

 \textbf{RANs-50} & \textbf{45.8M} & \textbf{0.87G} & \textbf{4.02ms} & \textbf{62.40} \\ \cline{1-5} 
                  
ResNet-101 & 149M & 2.29G & 6.86ms & 60.62 \\ \cline{1-5} 

\textbf{RANs-101} & \textbf{85.1M} & \textbf{1.52G} & \textbf{7.19ms} & \textbf{64.18} \\ \cline{1-5} 

ResNet-152 & 204M & 3.41G & 9.29ms & 61.57 \\ \cline{1-5} 

\textbf{RANs-152} & \textbf{117M} & \textbf{2.18G} & \textbf{9.72ms} & \textbf{66.16} \\ \hline
\end{tabular}
\caption{Image classification performance on the Tiny ImageNet datasets for $26, 35, 50, 101,$ and $152$-layer architectures.}
\label{tab_resultImageNet}
\end{table}

\subsubsection{MobileNet Architectures}
RAN-based MobileNet
where the 2D convolution layers replace by the RAN block, and the original MobileNetV1 show the direct effect of the RAN block in mobile-based shallower architectures. The RAN-based MobileNet performs more than $1\%$ in validation accuracy with $75$\% fewer trainable parameters and almost $67\%$ fewer FLOPs shown in Table \ref{tab_resultCifar}. Our RAN-based MobileNet takes similar latency to the original MobileNet.

\subsubsection{SqueezeNext Architectures}
We implement RAN-based SqueezeNext to show the effectiveness of the RAN block compared with the SqueezeNext block. Table \ref{tab_resultCifar} compares the performance for $23$-layer architecture with widening factors $1$ and $2$. Our RAN-based SqueezeNexts outperform the original SqueezeNexts with 34\%, 24\%, and 33\% fewer parameters, FLOPs, and latency, respectively.

\subsubsection{Recursive Networks}
Table \ref{tab_resultDRRNs} shows the Peak Signal-to-Noise Ratio (PSNR) results of several CNN models including DRRN, and our proposed RARNet on the Set5 dataset. The comparison between DRRN and RARNet is our main focus as it directly indicates the effectiveness of using our proposed RAN block. DRRN19 and DRRN125 are constructed using $B=1, U=9$, and $B=1, U=25$, respectively. For fair comparison, we also construct similar architecture like RARNet19 ($B=1, U_{RAN}=9$) and RARNet125 ($B=1, U_{RAN}=25$). Our proposed models outperform all CNN models in Table \ref{tab_resultDRRNs} on the Set5 dataset and for all scaling factors. As we propose a parameter-efficient architecture, parameter comparison is essential along with the testing performance. Our proposed model for $B=1$, and $U_{RAN}=9$ takes $18182$ parameters compared to $297,216$ parameters of DRRN ($B=1$, and $U=9$). RARNet, constructed using RAN blocks, reduces by $94\%$ the trainable parameters compared to the DRRN.

\begin{table}
\centering
\begin{tabular}{|l|c|c|}
\hline
\multirow{2}{*}{\shortstack{Model Name}} & \multicolumn{2}{c|}{Validation Accuracy} \\ \cline{2-3} 
& CIFAR10 & CIFAR100 \\ \hline
 WRN-28-10 \cite{zagoruyko2016wide} & 94.68 & 79.57\\ \cline{1-3}  
 WRAN-26-2 (Ours) & 96.32 & 83.54\\ \cline{1-3}
 WRAN-26-4 (Ours) &  96.68  & 83.75 \\ \cline{1-3}
 WRAN-26-6 (Ours) & 96.77 & 83.78 \\ \cline{1-3}
 WRAN-26-8 (Ours) & 96.83 & 83.82 \\ \cline{1-3}
 WRAN-26-10 (Ours) & 96.87 & 83.92 \\ \cline{1-3}\hline
\end{tabular}
\caption{Image classification performance on the CIFAR benchmarks for 26-layer architectures with different widening factors. }
\label{tab_resultWRNs}

\end{table}

\begin{table}[!ht]
\centering
\begin{tabular}{|l|c|c|c|} \hline
\multirow{2}{*}{\shortstack{Model Architecture}} & \multicolumn{3}{c|}{Scale} \\ \cline{2-4} & x2 & x3 & x4 \\ \hline
SRCNN \cite{dong2015image} & 36.66 & 32.75 & 30.48\\
VDSR \cite{kim2016accurate} & 37.53 & 33.66 & 31.35\\
DRCN \cite{kim2016deeply} & 37.63 & 33.82 & 31.53\\
DRRN19 \cite{tai2017image} & 37.66 & 33.93 & 31.58\\
DRRN125 \cite{tai2017image} & 37.74 & 34.03 & 31.68\\
RARNet19 & 37.73 & 33.99 & 31.63\\ 
RARNet25 & \textbf{37.84} & \textbf{34.11} & \textbf{31.84}\\\hline
\end{tabular}
\caption{Benchmark testing PSNR results for scaling factors $\times2$, $\times3$, and $\times4$ on Set5 dataset.} 
\label{tab_resultDRRNs}
\end{table}

\section{Discussion and Conclusions}
This work introduces a new block, the RAN block which is constructed with two sequential 1D DSCs and residual connections. This RAN block can be used as a replacement for any variation of 2D convolution. These modifications help to reduce trainable parameters, FLOPs, and latency, as well as improve validation performance on image classification tasks. 
We also checked this proposed block for widened ResNets, MobileNet, and SqueezeNext architectures and showed that the RANs-based wide ResNets, MobileNet, 
and SqueezeNext obtain better accuracy and take fewer parameters, FLOPs, and latency.  
We also checked the effectiveness of our RAN block on the SISR task to show how the RAN block performs for other areas than the classification. 
It was found that our proposed recursive axial ResNets (RARNets) improve image resolution and reduce around $94\%$ trainable parameters than the other CNN-based super-resolution models. Extensive experiments and analysis show that RANs can be shallow, deep, and wide. These are parameter-efficient and superior models for image classification and SISR. 
We have shown that our proposed model is a viable replacement for any 2D convolutional layer on the tested tasks. 
Further work is needed to determine the range of applications for which RANs may offer advantages.

{\small
\bibliographystyle{ieee_fullname}
\bibliography{RAN}
}

\end{document}